# Chimpanzee voice prints? Insights from transfer learning experiments from human voices

Maël Leroux[a,b,c*△], Orestes Gutierrez Al-Khudhairy[d△],

Nicolas Perony[d†], Simon W. Townsend[a,c,e†]

[a] Department of Comparative Language Science, University of Zurich, Switzerland

[b] Budongo Conservation Field Station, Masindi, Uganda

[c] Center for the Interdisciplinary Study of Language Evolution (ISLE), University of Zurich, Switzerland

[d] OTO.ai, New York City, USA

[e] Department of Psychology, University of Warwick, UK

△ the authors contributed equally and share first authorship

† the authors contributed equally and share last authorship

*Corresponding author:

M. Leroux, Department of Comparative Language Science, University of Zurich, Thurgauerstrasse 30/32, 8050 Zurich, Switzerland

E-mail Address: mael.leroux@uzh.ch

**Abstract**

Individual vocal differences are ubiquitous in the animal kingdom. In humans, these differences pervade the entire vocal repertoire and constitute a "voice print". Apes, our closest-living relatives, possess individual signatures within specific call types, but the potential for a unique voice print has been little investigated. This is partially attributed to the limitations associated with extracting meaningful features from small data sets. Advances in machine learning have highlighted an alternative to traditional acoustic features, namely pre-trained learnt extractors. Here, we present an approach building on these developments: leveraging a feature extractor based on a deep neural network trained on over 10,000 human voice prints to provide an informative space over which we identify chimpanzee voice prints. We compare our results with those obtained by using traditional acoustic features and discuss the benefits of our methodology and the significance of our findings for the identification of "voice prints" in non-human animals.

**Introduction**

The capacity to recognise one conspecific from another is well documented in non-human animals (hereafter animals) and is argued to be essential for social interactions [1]. The vocal modality has received particular interest in this regard [2]. Indeed, over the last 50 years, a growing body of evidence has highlighted the encoding of ID information within calls of a wide variety of taxa (amphibians [3], birds [4], mammals including non-primates [5] and primates [6]; for a review: see [2]) with a variety of potential adaptive social functions, ranging from kin cooperation [7], reinforcing bonding with conspecifics [8], reducing aggression [9] and facilitating social interactions in low-visibility, dense habitats, for example [10]. This, in part, has led to the hypothesis that sociality is an important evolutionary driver for the emergence of individual signatures (ID) encoded via vocal signals in animals [11]. Key support for this hypothesis derives from humans who arguably reside in the most complex of social systems [12] and also possess individual signatures. However, an important distinction between humans and animals is that, in humans, these differences pervade the entire vocal repertoire and can be better framed as an individually distinctive voice or "voice print". Interestingly, whether other animals are also capable of reliably encoding a unique ID signature across their repertoire remains little investigated, despite the potential evolutionary implications of such data.

Our closest living relatives, chimpanzees, have, like humans, also been shown to reside within complex multi-male multi-female social groups, composed of individuals from various age classes interacting with each other repeatedly and forming long-lasting bonds based on fission-fusion dynamics [13]. Critically, chimpanzees rely heavily on vocalisations to negotiate their complex social worlds. For example, pant-grunt greeting calls in chimpanzees are produced by subordinates to higher ranking individuals to signal submission [13] and can be flexibly produced with individuals inhibiting the production of pant-grunts to a higher ranking individual if the alpha male is present [14]. Pant-hoot calls on the other hand are long distance compound contact calls that allow individuals to keep track of other group member's positions [13]. Chimpanzees also regularly synchronise their pant-hoots, forming choruses, and it has been demonstrated that chorusing reflects bonding strength between the individuals calling, promoting affiliative behaviour [8,15]. Given chimpanzees' complex social dynamics as well as the dense nature of their forest habitat which frequently precludes visual contact, we hypothesised that a unique ID signature may be recurrently encoded in vocalisations across their repertoire, similar to a human voice print. Whilst previous work has identified individual signature in their calls (copulation call [16]; pant-hoot [17]; scream and whimper [18]), to our knowledge, no work has investigated the potential for a unique individual signature across call types in chimpanzees or great apes in general.

One explanation for this gap is the scarcity of appropriate methods currently at our disposal to explore such questions. Indeed, to date, animal communication researchers and acousticians have primarily relied on traditional acoustic analyses, extracting, often, arbitrarily chosen acoustic features from calls to highlight individual variation within the acoustic structure of a specific call and sometimes even multiple calls (bonobos [19]) for a given species (e.g., birds [20]; marine mammals [21], non-human primates [22]). Unfortunately, this approach to investigate individual differences in call(s) is confounded by the fact that any individual variation across different call types will be overpowered by the differences between the call types themselves. Recent developments in machine learning (ML) methods represent a potential route out of this impasse.

For the specific case of bioacoustics, machine learning algorithms have been applied to investigate ID encoding in animal vocalisations in both birds and mammals. In the case of birds, supervised classifiers have been applied to investigate ID signatures across the vocal repertoire of zebra finches [23]. The algorithms were able to automatically classify each call type according to the ID of the caller, and playback experiments confirmed the birds were able to associate ID signatures across call types [23]. However, the performance of the classifier dropped drastically when being trained and tested on different call types, suggesting zebra finches do not possess a unique vocal signature shared across their repertoire but rather an ID signature specific to each call type [23]. One potential reason for the lack of generalisability is the use of predefined and hand-curated acoustic features on which to perform machine learning classification of ID encoding (such as pitch, F0 etc.). This potentially limits the performance of the downstream classification task as strong prior assumptions as to which acoustic phenomena are important for characterising identity are made. In the case of mammals, Reby et al [24] investigated cross-call ID signatures in red deer. Here, acoustic feature extraction was performed using Mel-frequency cepstral coefficients (MFCCs) and subsequently, models trained to classify roar calls according to the signaller ID achieved an accuracy of 93.4% [24]. To test whether the model could generalise ID signatures in roars to other call types, a second model was trained with roars as the training set (90%) and random other call types as a test set (10%). This model classification accuracy dropped to 63.4% and while performing above chance, considerable individual and call type variation affected classification accuracy [24]. One possible cause for this drop in performance and the substantial variation is the unbalanced nature of the data set implemented. An alternative explanation could lie in the use of MFCCs. Indeed, the proper calibration of MFCCs becomes computationally inefficient as the number of tunable hyperparameters increases, and the complexity of the hyperparameter space results in "weak performance in the presence of noise" [25].

The field of machine learning has since progressed to learn how to optimise this balance, thus moving from static feature sets, derived through heuristics and intuition such as MFCCs, to using either completely learnt filters [26] or such filters in combination with predefined acoustic feature extractors [27,28]. Learnt filters are formed by large and deep neural networks (typically convolutional) that use fewer embedded prior information to learn efficient representations on large datasets, potentially leading to a less biased and more efficient feature set on which to perform subsequent machine learning tasks. Indeed, these filters have proven successful and increased the performance of automated classifications relative to historical feature extraction methods and they have been shown to exhibit state-of-the-art performance on a wide range of audio classification tasks, including animal sounds [26,27], especially in small-data regimes. Specifically, when comparing a fully trainable acoustic feature extractor (LEAF) to MFCCs, the learnt filters outperformed MFCCs, with the LEAF framework being seemingly able to focus on more informative frequency ranges for representing the underlying audio events than MFCCs [26]. This suggests the use of learnt features may allow for more informative representations than fixed feature extractors (e.g., MFCCs), if the former have been optimised appropriately relative to the classification task and acoustic phenomena.

In this paper, we present an approach that builds on these recent developments, whilst further investigating the potential for animal voice signatures. We achieved this by mitigating the most limiting issues faced when training acoustic feature extractors, i.e., the amount of data and computing infrastructure that is usually necessary to optimally tune them. Given feature extractors have a number of trainable parameters in the order of magnitude of thousands and sometimes millions [26–28], training them is especially problematic when working with animal datasets, which are often limited in their number of samples. To tackle this issue, we adopted an approach similar to that of [26], where the authors performed machine learning tasks using learnt feature extractors that had been *pre-trained* on a different set of machine learning tasks. Here, we aim to apply and investigate the benefits of such methodology by implementing DeepTone's (a neural network based architecture) *Identity* model as a feature extractor instead of directly training the frontend of our machine learning framework (see methods section). DeepTone's *Identity* model is a convolutional deep neural network (TCN) that has been trained to disambiguate human voice print on over 10,000 unique utterances. To investistigate call type independent voice signatures in chimpanzees, we provided the model with three different vocalisation elements extracted from three different individuals. The output of the *Identity* model is then expected to create an informative feature space over which we trained shallow classification models to identify chimpanzee voice prints. We directly compared our approach with more traditional feature extractors, MFCCs. Our prediction is that given the low data training regime, the use of a feature extractor that has been pre-trained on an extensive and varied data set may perform better than static feature sets such as MFCCs.

**Results**

The number of call types per chimpanzee whose acoustic features were extracted using the DeepTone *Identity* model are presented in Table 1.

Table 1 - Counts of call type (table columns) data points for each chimpanzee individual (table rows)

|  | Scream | Pant-hoot-intro | Pant-hoot-climax | All call types |
|---|---|---|---|---|
| Squibs (male, 28yo) | 30 | 29 | 13 | 72 |
| Nambi (female, 57yo) | 30 | 20 | 17 | 67 |
| Zed (male, 18yo) | 33 | 12 | 11 | 56 |
| Total | 93 | 61 | 41 | 195 |

*Identity signature in chimpanzee across call types*

DeepTone's *Identity* model was able to generate a feature space for which the identity of chimpanzees, irrespective of call types, can be reliably classified (above 80% accuracy for all classifiers tested, see Table 2). Specifically, Support Vector Machines (SVMs) had, on average, the highest F1 (i.e., a weighted average of recall and precision, 0.906 $\mp$ 0.002) and accuracy scores (i.e., the fraction of correctly predicted IDs 0.910 $\mp$ 0.001), whilst maintaining the least variation around mean performance. As such we found SVMs were the best classifiers out of those currently tested for identifying chimpanzee ID within this specific data set.

**Table 2** - Mean and standard errors of classification scores (table rows) for four machine learning models (table columns) used to classify chimpanzee ID within the calls described in Table 1. These summarising statistics are calculated each from a sample of 500 data points, which result from the random partition of training and test sets. The highest scores across classifiers for each metric are presented in bold font and belong to the SVM. SVM: Support Vector Machines, RF: Random Forests, NB: Naïve Bayes, GP: Gaussian Processes.

|  | SVM | RF | NB | GP |
|---|---|---|---|---|
| F1-score | **0.906 ∓ 0.002** | 0.864 ∓ 0.002 | 0.806 ∓ 0.003 | 0.881 ∓ 0.002 |
| Accuracy | **0.910 ∓ 0.001** | 0.869 ∓ 0.002 | 0.814 ∓ 0.003 | 0.887 ∓ 0.002 |

It is important to note the dataset was, to a certain degree, unbalanced given the unequal number of samples per call type and individual, and the presence of two males for only one female (Table 1). To ensure that our analysis was not biased by the sex of the individuals, we investigated the confusion matrices of the classification experiments (which provide detailed classification accuracy for each individual and call type, see Figure 1). Specifically, for both Zed and Squibs (males, Figures 1b & 1c), the highest proportion of misclassified pant-hoot-intro and pant-hoot-climax calls were due to confusion with Nambi (female), excluding sex as a source of bias. In terms of call type, for it to be a source of bias and a confounding factor, we would expect each chimpanzee to have a specific call type or subset that is different for each individual, and significantly more accurate in predicting their ID, than other call types. We see that this is not the case because all individuals consistently share a call type (screams) that can predict ID with an accuracy of over 90%, and which is also the most frequently appearing and balanced call type per chimpanzee (see Table 1).

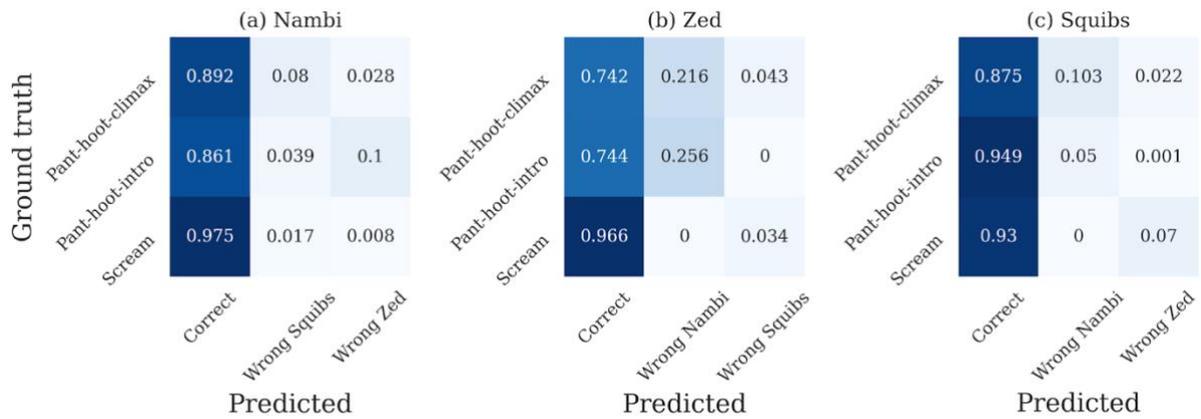

**Figure 1** - Confusion matrices depicting the proportion of times that the call type corresponding to each row was either correctly or incorrectly identified as belonging to the chimpanzee that uttered it. The title of each confusion matrix informs on the particular chimpanzee individual, and the correspondence of rows with the left column, "Correct", informs on how many times each call was correctly identified as belonging to said chimpanzee. Instead, the correspondence of rows with middle & right columns informs on the proportion of times that a particular call type was confused with another chimpanzee individual. For example, confusion matrix '(c) Squibs' tells us that Squibs' *Pant-hoot-climax* calls were correctly identified 87.5% of times, and that they were incorrectly identified as Nambi's and Zed's 10.3% and 2.2% of the times respectively. The proportions are calculated relative to the total number of each call type that belongs to a chimpanzee, such that summing each row should equal one.

*Comparison with MFCCs*

To assess the potential advantage of leant feature extractors compared to the current state-of-the-art static MFCCs, we compared classification accuracy of SVMs using either the DeepTone *Identity* model or MFCCs.

DeepTone's average accuracy was consistently higher than MFCCs across the whole range of training data points used (see Figure 2). Critically, the advantage of DeepTone was maximal for smaller ranges of the number of training points used (around the 40-sample mark).

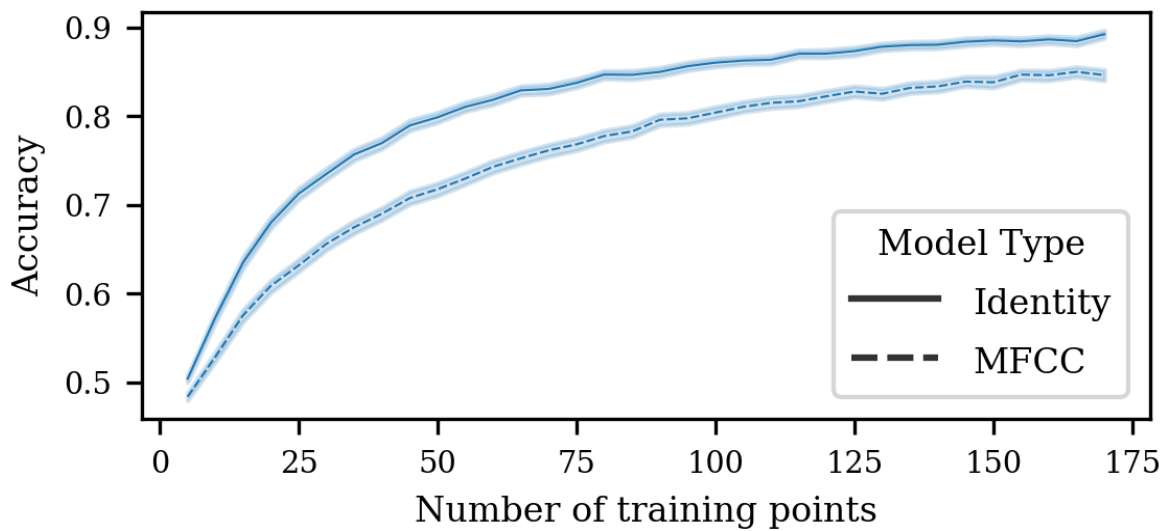

**Figure 2** - Accuracy as a function of the fraction of data used for training. Each point represents the average of a sample constructed using 2000 replicates, which are derived by sampling the margins parameter "C" 500 times respectively from the four following bin ranges (0.1-1, 1-10, 10-100, 100-1000). The blue shaded curves represent the 95% confidence interval of the estimated averages. The line types correspond to the type of feature extractor used: the Identity model (solid) or the MFCC (dashed).

**Discussion**

Using a state-of-the-art machine learning algorithm, we show that chimpanzees, our closest living relatives, possess a unique individual vocal signature that is shared across call types tested in this study: namely screams and pant hoot call elements. These results suggest ID encoding in chimpanzee vocal communication is independent of the nature of the vocalisation, similarly to a voice print in humans. Specifically, implementing a temporal convolutional deep neural network previously trained on human voice (DeepTone's *Identity* model), we successfully extracted acoustic features and classified three distinct vocal units from the chimpanzee repertoire based on the caller's identity. We compared four different classifiers and demonstrated the greatest accuracy when using support vector machines, further validating their current use in the field of animal bioacoustics [17]. Critically, ID classification was not attributed to a specific call type for each individual, confirming classification accuracy was not biased by call types. Furthermore, given most incorrect classifications for males were attributed to the female and not the other male, we also ruled out sex as a confounding factor in the classification accuracy. In addition, the number of calls per individual did not explain classification accuracy, suggesting the sensitivity of the model to the amount of data available is limited. Finally, we show that DeepTone is a better feature extractor compared to MFCCs for the classification of vocal signatures, providing the greatest advantage over smaller ranges of training points.

Together, this work establishes DeepTone's learnt feature extractors, the *Identity* model, in conjunction with a shallow classifier, as the most robust method for characterizing chimpanzee ID voice signatures within this dataset. To our knowledge, we provide the first robust evidence for call type independent ID voice signatures in animals. This, in turn, suggests that individually distinctive voice (as opposed to call) structuring is unlikely to be unique to humans and might be a more general feature of animal communication systems. Indeed the communication of ID, independent of call type, is likely to aid the successful navigation of numerous interactional challenges often encountered by social species, such as chimpanzees. Moreover, from a perceptual or cognitive perspective, it would be more parsimonious to monitor individuals from their voice as opposed to learning individual signatures for each call type.

Our findings also provide promising insight into the potential generalisation of ML-algorithms to non-human systems. Specifically, despite our comparatively small sample size, we could still successfully classify calls according to ID. We are therefore confident that further computational developments will facilitate the addressing of identical questions in other species. For instance, extensive work has already highlighted the encoding of ID within tonal vocalisations comprising the bonobo repertoire [19]. However, whether the ID encoding was distinct for each vocalisation or common across the repertoire remains unknown. Bonobos, the other species from the Pan genera, are with chimpanzees, our closest living relatives [29]. As such, evidence for a unique voice signature in this species is central to unpacking the fine-grained evolution of this phenomenon in humans.

The successful generalisations of DeepTone to the chimpanzee system, albeit previously trained on human voice, further tentatively highlights the similarity between human and chimpanzee acoustic systems. In humans, it has been shown that individual voice recognition relies on the phonological structure of a speaker's voice [30,31]. Phonemes in human language are meaningless sounds combined together to form meaningful units (e.g., words) which in turn, can be syntactically combined into higher-order structures (i.e., phrases) [32,33]. Ultimately, evidence that voice recognition in humans operates at the phonological level means that phonemes, irrespective of higher-order structuring - i.e., speech content - are enough for the recognition of an individual voice. A key extension to further expand our understanding of a potential continuity between human and chimpanzee voice prints would be to investigate whether similar mechanisms are responsible for an individual voice signature in chimpanzees. However, while distinct call types can be considered as homologues to words (as they are meaning-bearing units [34]), no work has investigated the potential for a phonological-like structuring in chimpanzees - i.e., whether call type structures are combinations of recurring meaningless units together. Such data would be central to support the claim that similar mechanisms underlie voice recognition in chimpanzees and we hope future research will address this question.

In conclusion, using a novel machine learning approach that transfers learning from human speech and is capable of generalising to non-human primate systems, we provide the first evidence for a voice print-like capacity in our closest-living relatives. How chimpanzees perceive these voice prints specifically or which mechanisms underlie such cognitive processes remains to be investigated but minimally, our findings represent a promising step forward in the objective quantification of acoustic variation in animal calls.

**Methods**

*Study site and data collection*

The data for this study were collected at the Budongo Conservation Field Station (BCFS), Budongo forest, Uganda. The Sonso community consisted of 72 individuals at the time of the study, including 43 adults (12 males, 31 females), 11 subadults (2 males, 9 females), 10 juveniles (8 males, 2 females) and 8 infants (4 males, 4 females).

The data were collected over an 18-month period (April-August 2018, February-June 2019 and September 2019-March 2020). We followed adult individuals using focal animal sampling for a duration of 2 hours at a time, totalling 361 focal hours. During a focal follow, all vocalisations were recorded using a Marantz PMD661 mk3 handheld digital audio recorder (Marantz, Japan, sample rate 44.1kHz, resolution 32bits, uncompressed PCM format) connected to a directional Sennheiser ME66/K6 microphone (Sennheiser, Germany). For each vocalisation recorded, the caller ID and call type was noted after the calls. In addition, *ad libitum* sampling was used to record every vocalisation for which the caller could be identified visually.

*Data processing*

Call bouts were extracted using Praat, Adobe Audition and Audacity. Manual call type classification was based on [35] via spectrogram inspection. To ensure the analyses reach a strong enough power, we selected call bouts from only 2 vocalisations produced by 3 individuals for which we could gather a substantial data set, namely screams and pant hoots, produced by one adult female (Nambi), and two adult males (Squibs and Zed). To avoid biasing the classification process and ensure the algorithm classified chimpanzee vocal production and not noise or sounds coming from the surrounding environment, we further segmented each call bout, excluding segments that did not contain any chimpanzee signals and elements in which noise overtook the signals. Elements were defined as a continuous F0 trace and are similar throughout a scream bout. However, the pant-hoot is a compound call composed of four distinct phases: introduction, build-up, climax and let-down [35] and elements' acoustic properties differ across these phases. Because build-up and let-down elements were not often clearly identifiable, and because previous research has shown introduction and climax phases are responsible for ID encoding in the pant-hoot [17], we only selected elements from the latter two phases - i.e., introduction and climax. Therefore, overall, we integrated 3 distinct elements for 3 different individuals: scream, pant-hoot-introduction and pant-hoot-climax elements (see Figure 3).

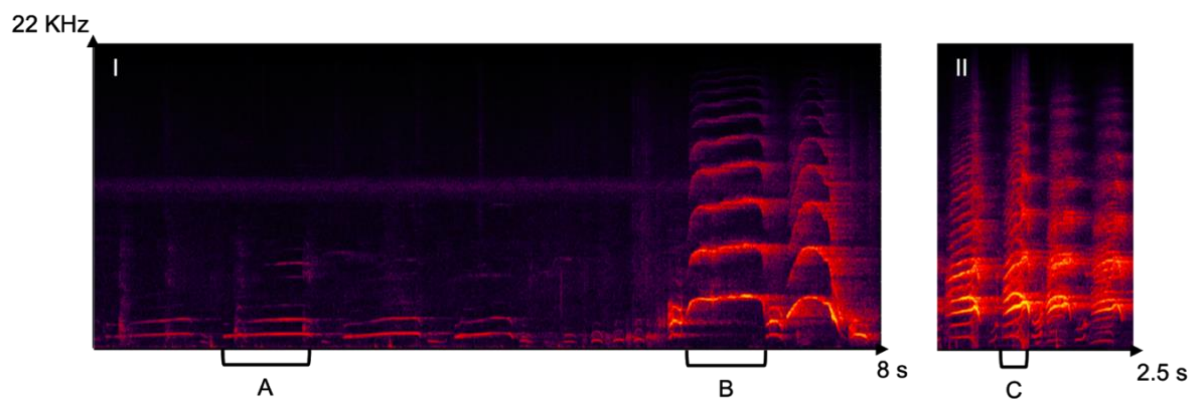

**Figure 3** - Spectrograms of a pant-hoot (I) and a scream bout (II). Elements from the introduction (A) and climax (B) phases of the pant-hoot and from the scream bout (C) were extracted.

*DeepTone's model architecture and the resulting chimpanzee voice print embeddings space*

DeepTone is a deep neural network architecture that takes the amplitude time-series of waveforms as an input. This is first passed through a series of residual temporal convolutional (TCNRes) blocks, which have been shown to outperform certain counterparts, including recurrent neural networks, on a variety of machine learning tasks [36], and optimises roughly an order of magnitude less parameters relative to typical *transformer* architectures used for speech detection (e.g., [37] requires the optimisation of 85.4 million parameters). The output of the TCNRes blocks is then passed through a one-dimensional convolutional (Conv1D) filtering layer. The final output is a compressed latent representation of the much denser waveform, an *Identity* vector, **I**, of 128 dimensions that is generated for every 64 milliseconds of audio. For an audio snippet of arbitrary length, DeepTone will therefore create an embedding space (i.e., sets of *Identity* vectors belonging to a particular individual) with dimension size relative to the length of the audio. To accommodate the variable audio-snippet lengths (which would subsequently result in a variable number of dimensions for each audio snippet's time-resolved embedding) we performed mean pooling over the embedding space, i.e., a time average over the identity vectors outputted per audio snippet. The resultant data set used for classification, which is formed by 195 vectors (one per audio snippet), each of 128 dimensions, is outputted for every 64 milliseconds of inputted audio sampled at 16 kHz.

The weighting of interactions that form the core calculations of TCNRes and Conv1D blocks, referred to as the *parameterisation*, are optimised during training (using samples from 10,268 human speakers split between train, validation, and test sets, with 59 samples per speaker on average), relative to the type of loss function and optimisation algorithm used. The *Identity* model implements a triplet loss function, and therefore attempts to simultaneously minimise the Euclidean distance of the latent representation of utterances belonging to the same speaker, whilst maximising the distance between utterances belonging to different speakers. In this way, DeepTone does not learn the identity of particular speakers, instead, it learns which interactions within the amplitude time-series of the waveform (controlled by the parameters of the TCNRes) and patterns within these interactions (detected by the Conv1D block), are important for generating features (the 128 dimensions of each *Identity* vector), that can maximally separate clusters of class-interspecific acoustic identity in the resulting embedding space. To actually classify the identity of each outputted *Identity* vector, a classifier has to be trained to identify chimpanzee voice signatures within DeepTone's embedding space.

*Identifying optimal shallow supervised machine learning classifiers*

We trained four frequently used classifiers (Support Vector Machines, Random Forests, Naive Bayes and Gaussian Processes) to verify that the identity of chimpanzee individuals is robustly represented in the DeepTone features space. We ran each classifier on 500 random partitions of the training and test set (75% train / 25% test), and calculated the classification scores of each random partition for each classifier.

*Comparison of DeepTone with MFCCs*

Once an optimal combination of the DeepTone feature space and classifier was found we subsequently used it to compare DeepTone to the MFCC feature space. The comparison consisted in running multiple classification experiments, pairing the SVM classifier with the respective MFCC and DeepTone features spaces. During the experiments, we (i) modulated the number of training points over a fixed range (ranging from 5 to 175 points) and (ii) sampled the SVMs margin hyperparameter "C" within four, fixed, non-overlapping bins, 500 times for each bin, with an exponentially increasing bin-range size (0.1-1, 1-10, 10-100, 100-1000) to compare the classification capabilities of DeepTone and MFCCs.

The set-up for comparing MFCCs to DeepTone was chosen by anticipating the effect of the possible hyperparameters of the experiment on the outcome. Specifically we sought to find a sufficient and significant set of hyperparameters and metrics to iterate over, relative to the core principles of the comparison (which we explain below). Similarly we aimed to understand which hyperparameters we should fix to ensure a comparable experimental set up between the feature spaces being compared.

1) Core principles of the comparison

The core principle is to compare the ability for each feature space to generate information that is useful for generally (i.e., not specific to an individual or call type) capturing and subsequently classifying chimpanzee voice signatures independently of call type. This means we wished to know which feature space is better at creating label specific clusters of data points that can be distinguished, relative to the classifier used.

2) Metric for measuring the core principle of comparison and its implication

Given that accuracy is a metric that has been used in a similar study [17] we decided to also use it. Specifically, accuracy measures the ratio of correct predictions over the total number of predictions. Given the type of classifier we were using, accuracy technically informs on the ability for the SVMs to find the most optimal hyperplanes for separating clusters of data. Thus for identical data sets, the feature space with the best accuracy directly implies that it creates clusters with the least amount of overlap between the clusters of classes we wish to separate.

3) Fixed hyperparameters

We decided to fix the number of cepstral features of the MFCC to equal the number of dimensions in DeepTone's feature space. This choice is driven by the well known influence that the number of dimensions (for any type of feature used to inform on class types) has on the performance of any type of classifier that is used (known as the *curse of dimensionality,* for more see the review by [38]).

We fixed the minimum and maximum frequencies of the MFCC to mitigate the possibility of uninformative frequency diluting the cepstral features. We set the minimum frequency to 50 Hz, which is a standard choice for avoiding potential sources of noise [39–41]. The maximum was chosen according to Nyquist–Shannon sampling theorem (also known as Nyquist's frequency principle), which states the theoretically observed maximum frequency is roughly half the sample rate of the amplitude time-series of audio. This limits the number of potentially uninformative MFCC features, which is also implemented by [41]. Accordingly, given the native sample rate of the data is 44.1 kHz, the maximum frequency allowed for the MFCCs was set to 22.05 kHz.

4) Variable hyperparameters

The number of training points directly affects the performance of the classifier, and should directly show how informative the two feature spaces are by assuming more data directly translates into more potential information. Subsequently, if one feature space has achieved higher accuracy, independently of the number of training points, this implies it is able to generate tighter and more separable clusters of class specific data, and therefore more meaningful representations of vocal identity, which as stated earlier, forms the core principle of the feature space comparison. Therefore, we compared how well each classifier performed relative to different amounts of information provided by varying the number of training points.

Finally, we also decided to vary the SVM margin hyperparameter "C". This parameter controls how sensitive the loss function is to the miss-labeled occurrences during training, by allowing for more/less miss labelling of the training data. This parameter therefore controls the amount of overfitting to the training data (i.e., when a classifier although learning how to classify the training data well, is not able to do so for the test data). The "C" parameter therefore also controls the final accuracy score of training and test data, on which the principal of our comparison hinges on. Because optimisation of "C" is usually achieved relative to the accuracy of the test data, this implies it is overfitted to the specific data set used, i.e., the call types and chimpanzee individuals used in our experiment. Given these are not necessarily reflective of all possible call types that all chimpanzees may produce, and that the core principle of the feature space comparison is to understand how well each feature space represents the general chimpanzee vocal repertoire (rather than an individual and call specific one), we chose not to optimise "C". Instead, we decided to report the average accuracy scores for a range of "C" in which it is likely that the "true" optimal value occurs, relative to the task of identifying voice signatures of all chimpanzees and across all call types.

**Acknowledgments**

We thank UWA, UNCST and the President's office for permission to collect data in the Budongo Conservation Field Station. We are grateful for the constant support provided by all BCFS staff and the Royal Zoological Society of Scotland (RZSS). We thank Pawel Fedurek and Guillaume Dezecache for valuable comments on previous version of the manuscript. This work was supported by the Swiss National Science Foundation (PP00P3_163850) to S.W.T. and the NCCR Evolving Language (Swiss National Science Foundation Agreement #51NF40_180888). We declare no conflict of interest.